\documentclass[10pt,journal,twoside,cspaper,compsoc]{IEEEtran}

\usepackage[colorlinks=true]{hyper ref}

\ifCLASSOPTIONcompsoc
\else
\fi

\ifCLASSINFOpdf
   \usepackage[pdftex]{graphicx}
   \graphicspath{{.}{fig_eps/}}
\else
\fi

\usepackage[cmex10]{amsmath}
\usepackage{amssymb}
\usepackage{amsthm}
\usepackage{nicefrac}
\usepackage[ruled]{algorithm2e}
\usepackage{fixme}
\usepackage[usenames,dvipsnames]{color}
\usepackage{nth}
\usepackage[usenames,table]{xcolor}
\usepackage{multirow}
\usepackage{booktabs}

\usepackage{tikz}                  \usetikzlibrary{calc,arrows,positioning,shapes,shadows,spy,snakes,plotmarks,matrix,fit,backgrounds}
\usepackage{pgfplots}

\usepackage[small,bf]{caption}

\usepackage{fixltx2e}
\usepackage{url}

\newcommand{\ie}{\textit{i.e.}}
\newcommand{\eg}{\textit{e.g.}}
\newcommand{\etal}{\textit{et al.}}
\newcommand{\cf}{\textit{cf.}}
\newcommand{\tr}{\text{tr~}}
\DeclareMathOperator{\myspan}{span}
\DeclareMathOperator{\mydiag}{diag}

\DeclareMathOperator{\rlog}{Log}

\newcommand{\rk}[1]{{\color{black}{#1}}}

\newcommand{\yi}[1]{{\color{black}{#1}}}
\newcommand{\mn}[1]{{\color{black}{#1}}}

\newenvironment{rcases}
  {\left.\begin{aligned}}
  {\end{aligned}\right\rbrace}

\begin{document}
\title{Parametric Regression on the Grassmannian}
\author{Yi Hong, Nikhil Singh, Roland Kwitt, Nuno Vasconcelos and Marc Niethammer
\IEEEcompsocitemizethanks{\IEEEcompsocthanksitem Y. Hong, N. Singh and M. Niethammer are
with the Department of Computer Science, University of North Carolina, Chapel Hill, NC, USA.
E-mail: yihong@cs.unc.edu, nsingh@cs.unc.edu, mn@cs.unc.edu
\IEEEcompsocthanksitem R. Kwitt is with the Department of Computer Science,
University of Salzburg, A-5020 Salzburg, Austria.
E-mail: \mn{rkwitt@gmx.at}
\IEEEcompsocthanksitem N. Vasconcelos is with the Department of Electrical and Computer Engineering, 
University of California San Diego, CA, USA. \protect\\
E-mail: nvasconcelos@ucsd.edu}
}

%
%

\markboth{}%
{Hong \MakeLowercase{\textit{et al.}}: Parametric Regression on the Grassmannian}

\IEEEcompsoctitleabstractindextext{%
\begin{abstract}
We address the problem of fitting parametric curves on the Grassmann manifold for the purpose
of intrinsic parametric regression. As customary in the literature, we start from the energy minimization 
formulation of linear least-squares in Euclidean spaces and generalize this concept to general 
nonflat Riemannian manifolds, following an \textit{optimal-control} point of view. We then specialize this idea
to the Grassmann manifold and demonstrate that it yields a simple, extensible 
and easy-to-implement solution to the parametric regression problem. In fact, it allows us to extend the basic 
geodesic model to (1) a ``time-warped'' variant and (2) cubic splines. We demonstrate the 
utility of the proposed solution on different vision problems, such as shape regression as a function of age,
traffic-speed estimation and crowd-counting from surveillance video clips. Most notably, these problems 
can be conveniently solved within the same framework without any specifically-tailored steps along the 
processing pipeline.
\end{abstract}

}

\maketitle

\IEEEdisplaynotcompsoctitleabstractindextext
\IEEEpeerreviewmaketitle

\section{Introduction}
\label{section:introduction}

\IEEEPARstart{M}{any} data objects in computer vision problems admit a subspace representation. 
Examples include feature sets obtained after dimensionality reduction via principal component 
analysis (PCA), observability matrix representations of linear dynamical systems, or 
landmark-based representations of shapes. Assuming equal dimensionality (\eg, the same 
number of landmarks), data objects can be interpreted as points on the Grassmannian 
$\mathcal{G}(p,n)$, \ie, the manifold of $p$-dimensional linear subspaces of $\mathbb{R}^n$. 
The seminal work of \cite{Edelman98a} and the introduction of efficient processing algorithms 
to manipulate points on the Grassmannian \cite{Gallivan03a} has led to a variety of principled 
approaches to solve different vision and learning problems. These include domain adaptation 
\cite{Gopalan11a,Zheng12a}, gesture recognition \cite{Lui12a}, face recognition under 
illumination changes \cite{Lui09}, or the classification of visual dynamic processes \cite{Turuga11a}. 
Other works have explored subspace estimation via conjugate gradient descent \cite{Mittal12a}, mean 
shift clustering \cite{Cetinguel09a}, or the definition of suitable kernel functions 
\cite{Hamm08a,Jayasumana14a,Harandi14b} that can be used with a variety of kernel-based machine 
learning techniques.

\begin{figure*}[t]
\centering{
  \includegraphics[width = 1.5\columnwidth]{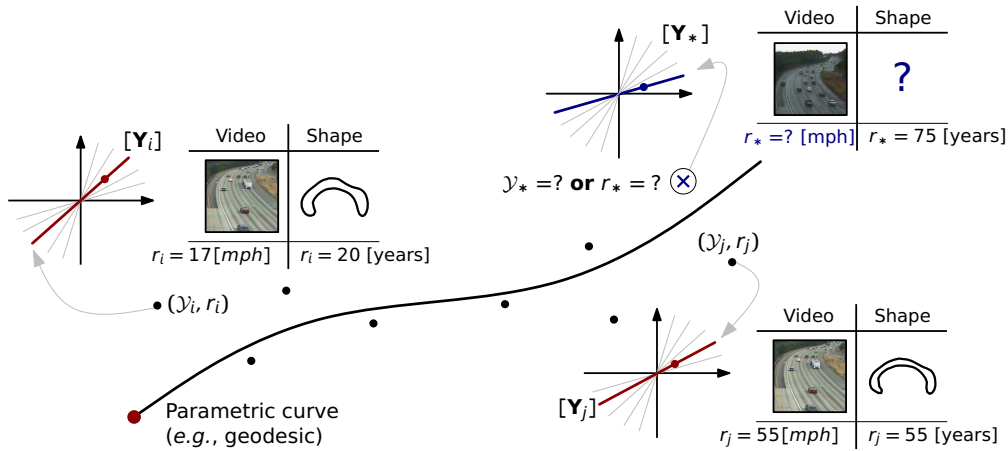}}
\caption{\label{fig:principle}
\small{Illustration of parametric regression and inference. At the point marked $\otimes$, 
the inference objective for (i) traffic videos is to predict the independent variable $r_{*}$ 
(\eg, speed), whereas for (ii) corpus callosum shapes we seek the manifold-valued $\mathcal{Y}_*$ 
at an independent variable (\eg, age). Here, elements on the 
Grassmannian are visualized as lines through the origin, \ie, $\mathcal{Y}_i \in \mathcal{G}(1,2)$.}}
\end{figure*}
 
Since, most of the time, the primary objective is to perform classification or recognition 
tasks on the Grassmannian, the problem of intrinsic regression in a parametric setting 
has gained little attention. However, modeling the relationship between manifold-valued
data and associated descriptive variables has the potential to address many problems 
in a principled way. For instance, it enables predictions of the descriptive variable while
respecting the geometry of the underlying space. Further, in scenarios 
such as shape regression \mn{--- a common problem in computational anatomy ---} we are 
specifically interested in summarizing continuous trajectories that capture variations in 
the manifold-valued variable as a function of the scalar independent variable. 
Fig. \ref{fig:principle} illustrates these two inference objectives. While predictions of the 
scalar-valued variable could, in principle, be formulated within existing frameworks 
such as Gaussian processes or support vector regression, \eg, by using Grassmann kernels 
\cite{Hamm08a,Jayasumana14a}, it is unclear how to or if it is possible to address 
the second inference objective in such a formulation.

In this work, we propose an approach to intrinsic regression that allows us to directly 
fit parametric curves to a collection of data points on the Grassmann manifold, indexed 
by a scalar-valued variable. Preliminary versions of this manuscript \cite{Hong2014a,Hong2014b}
essentially focused on fitting geodesics and how to re-parametrize the independent variable 
to increase flexibility. Here, we first recapitulate the \textit{optimal-control} perspective of 
curve fitting \mn{in Euclidean space as an example} and then discuss \mn{extensions of linear and cubic spline regression} on the Grassmannian. \mn{The} proposed models are \textit{simple} and 
natural extensions of classic regression models in Euclidean space. They provide a 
\textit{compact} representation of the complete curve, as opposed to discrete curve fitting 
approaches for instance which typically return a sampling of the sought-for curves. In
addition, the parametric form of the curves, \eg, given by initial conditions, allows to 
freely move along them and synthesize additional observations. Finally, parametric 
regression opens up the possibility of statistical analysis of curves on the manifold,
\mn{which is essential} for comparative studies in medical imaging for instance.

We demonstrate the versatility of the approach on two types of vision problems where 
data objects admit a representation on the Grassmannian. First, we model the aging 
trends in human brain structures and \mn{the} rat calvarium under an affine-invariant 
representation of shape \cite{Begelfor06a}. Second, we use our models to predict traffic speed 
and crowd counts from dynamical system representations of surveillance video clips 
\textit{without} any specifically tailored preprocessing. All these problems 
are solved \rk{within the same} framework with minor parameter adjustments.

The paper is organized as follows: Section \ref{sec:previouswork} discusses 
related work about regression on nonflat Riemannian 
manifolds. Section \ref{section:linear_regression} recapitulates the 
problems of linear, time-warped and cubic spline regression 
in Euclidean space from an optimal-control point of view. These ideas are then
extended to Riemannian manifolds (Section \ref{section:geodesic_regression}) 
and specialized to the Grassmannian (Section \ref{section:grassmannian_geodesic_regression}).
Experiments on toy examples and real applications
are presented in Section \ref{section:experiments_on_toy_examples} and 
\ref{section:experiments}, respectively. Section \ref{section:conclusion} 
concludes the paper with a review of the main points and a discussion of open problems.



\section{Previous work}
\label{sec:previouswork}

At the coarsest level, we distinguish between two categories of regression 
approaches: \textit{parametric} and \textit{non-parametric} strategies, with all the 
known trade-offs on both sides \cite{Moussa98a}. In fact, non-parametric 
regression on \textit{nonflat} manifolds has gained considerable attention
over the last years. Strategies range from kernel regression \cite{Davis07a} 
on the manifold of diffeomorphic transformations to gradient-descent \cite{Samir12a} 
approaches \mn{on manifolds} commonly encountered in computer 
vision \cite{Su12a}, such as the group of rotations $SO(3)$ or Kendall's shape space.
In other works, discretizations of the curve fitting problem have been explored
\cite{Boumal11a,Boumal11b} which, in some cases, even allow to employ
second-order optimization methods \cite{Boumal13a}. 
Because our work is a representative of the \textit{parametric} category, we 
mostly focus on parametric approaches in the following review. 

While differential geometric concepts, such as geodesics and intrinsic higher-order curves, 
have been well studied~\cite{noakes1989cubic, camarinha1995splines},
their use for parametric regression, \ie, finding parametric relationships between the
manifold-valued variable and \rk{an} independent scalar-valued variable, has 
only recently gained interest. 
A variety of methods extending concepts of regression 
in Euclidean spaces to nonflat manifolds have been proposed.
Rentmeesters~\cite{quentin2011}, Fletcher~\cite{Fletcher13geodesic} and 
Hinkle \etal~\cite{hinkle2014} address the problem of geodesic fitting on 
Riemannian manifolds, primarily focusing on symmetric spaces, \mn{to which}
the Grassmannian belongs.
Niethammer \etal~\cite{Niethammer11a} generalized linear regression 
to the manifold of diffeomorphisms to model image time-series data, 
followed by works extending this concept \cite{hong2012,singh2013vm} 
and enabling the use of higher-order models \cite{singh2014}.

From a conceptual point of view, we can identify two groups of  
solution strategies to solve \rk{parametric} regression problems on nonflat manifolds: first, \textit{geodesic shooting} 
based strategies which \mn{address} the problem using adjoint methods from an optimal-control 
point of view~\cite{Niethammer11a,hong2012,singh2013vm,singh2014}; the second 
group comprises strategies which are based on optimization techniques that leverage 
\textit{Jacobi fields} to compute the required gradients~\cite{quentin2011,Fletcher13geodesic}. 
Unlike Jacobi field approaches, solutions using adjoint methods do not require
computation of the curvature explicitly and easily extend to higher-order models, 
\eg, polynomials~\cite{hinkle2014} or splines~\cite{singh2014}. Our approach is 
a representative of the \mn{adjoint approach, thereby ensuring} extensibility to more advanced models, 
such as the proposed cubic splines extension.

In the context of computer-vision problems, Lui \cite{Lui12a} recently adapted the known 
Euclidean least-squares solution to the Grassmann manifold. While this strategy works 
remarkably well for the presented gesture recognition tasks, the formulation does not guarantee 
\mn{the minimization of} the sum-of-squared geodesic distances within the manifold, which would be the natural extension of least-squares to Riemannian manifolds according to the regression literature. \mn{Hence, the geometric} and variational interpretation of \cite{Lui12a} remains unclear.
In contrast, we address the problem from the
aforementioned energy-minimization point of view which allows us to
guarantee, by design, the consistency with the geometry of the manifold.

To the best of our knowledge, the closest works to ours are \cite{batzies2015geometric},
 \cite{quentin2011} and, to some extent, \cite{hinkle2014}. Batzies \etal~\cite{batzies2015geometric} 
discuss only a theoretical characterization of the geodesic fitting problem on the 
Grassmannian, but do not provide a numerical strategy for estimation. 
In contrast, we derive alternative optimality conditions using principles from 
optimal-control. These optimality conditions not only form the basis 
for our shooting approach, but also naturally lead to a convenient iterative 
algorithm. By construction, the obtained solution is guaranteed to be a geodesic. As discussed above, 
Rentmeesters~\cite{quentin2011} follows the Jacobi field approach. While 
both optimization methods have the same computational complexity for the gradient, 
\ie,  $O(np^2)$ on the Grassmannian $\mathcal{G}(p,n)$, it is not trivial to generalize 
\cite{quentin2011} to higher-order models. Hinkle \etal~\cite{hinkle2014}
address the problem of fitting polynomials, but mostly focus on manifolds with a Lie 
group structure\footnote{$G(p,n)$ does not possess such a group structure.}. In that case, adjoint 
optimization is greatly simplified. \mn{However, in general, 
curvature computations are required which can be tedious.} Our approach, on the other 
hand, offers an alternative, simple solution that is (i) extensible, (ii) easy to implement
and (iii) does not require specific knowledge of differential geometric concepts such as curvature
or Jacobi fields.

\begin{figure}[t!]
	\begin{center}
	\includegraphics[width=0.95\columnwidth]{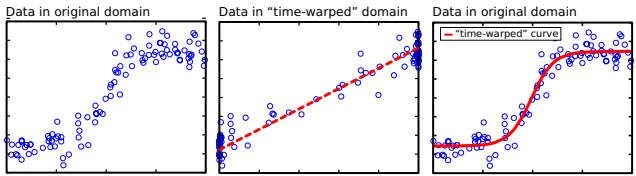}
	\end{center}
\caption{\label{fig:time_warped_least_square}\small{Illustration of time-warped regression in $\mathbb{R}$. The dashed straight-line (middle) shows the 
fitting result in the warped time coordinates, and the solid curve (right) demonstrates 
the fitting result to the original data points (left).}}
\end{figure}

\begin{figure*}[t]
	\begin{center}
	\includegraphics[width=0.95\textwidth]{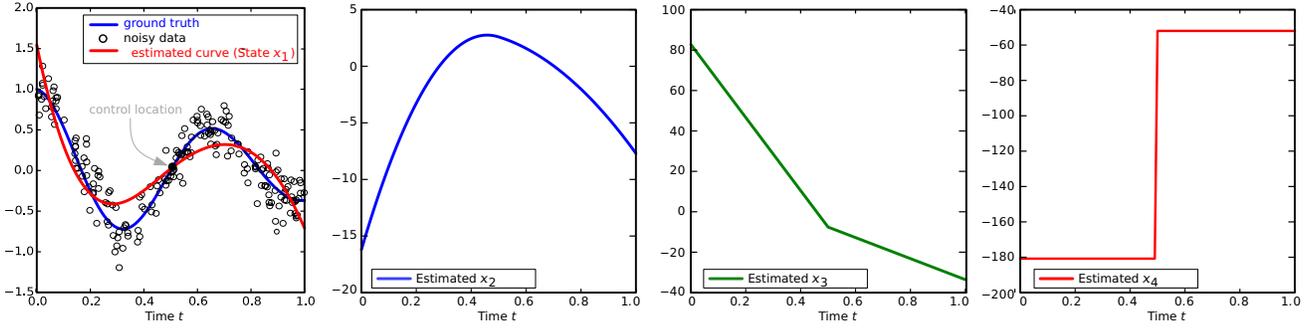}
	\end{center}
\caption{\label{fig:cublic_spline_euclidean}\small{Cubic spline regression in $\mathbb{R}$.
The leftmost side shows the regression result, and the remaining plots 
show the other states.}}
\end{figure*}

\section{Regression in $\mathbb{R}^n$ via Optimal-Control}
\label{section:linear_regression}

We begin our discussion with a review of linear regression in Euclidean space 
($\mathbb{R}^n$) and discuss its solution via optimal-control. While regression 
is a well studied statistical technique and several solutions exist for univariate 
and multivariate models, we will see that the presented 
optimal-control perspective not only allows to easily 
generalize regression to manifolds but also to define 
more complex parametric models on these manifolds.

\subsection{Linear regression}
\label{subsection:an_optimal_control_perspective}

A straight line in \rk{$\mathbb{R}^n$} can be defined as an
acceleration-free curve with parameter $t$, represented by states,
$(x_1(t), x_2(t))$, such that $\dot{x}_1 =x_2, \mbox{ and } \dot{x}_2
=0,$ where $x_1(t) \in \mathbb{R}^n$ is the position of a particle at
time $t$ and $x_2(t) \in \mathbb{R}^n$ represents its velocity at $t$.
Let $\{y_i\}_{i=0}^{N-1} \in \mathbb{R}^n$ denote a collection of $N$
measurements at time instances $t_0,\ldots,t_{N-1}$ with $t_i \in[0,1]$.  We define the linear
regression problem as that of estimating a parametrized linear motion
of the particle $x_1(t),$ such that the path of its trajectory best
fits the measurements in the least-squares sense. The unconstrained
optimization problem, from an optimal-control perspective, is
\begin{equation}
\begin{split}
\min_{\boldsymbol{\Theta}} & \quad E(\boldsymbol{\Theta}) = \sum_{i=0}^{N-1} \|x_1(t_i)-y_i\|^2 + \\
 & \quad \int_{0}^{1}\lambda_1^\top(\dot{x}_1-x_2) + \lambda_2^\top(\dot{x}_2)~dt \enspace,
\end{split}
\label{eqn:euclidan_linefit_control}
\end{equation}
with $\boldsymbol{\Theta} = \{x_i(0)\}_{i=1}^2$, \ie, the
\textit{initial conditions}, and $\lambda_1, \lambda_2 \in
\mathbb{R}^n$ are time-dependent Lagrangian multipliers.  
For readability, we have omitted the argument $t$ for $\lambda_1(t)$ and
$\lambda_2(t)$. These variables are also referred to as
\textit{adjoint} variables, \rk{enforcing} the dynamical
``straight-line'' constraints.  Evaluating the gradients with respect
to the state variables results in the \textit{adjoint system} as 
$
\dot{\lambda}_1 = 0, \mbox{ and } -\dot{\lambda}_2 =\lambda_1,
$
with jumps in $\lambda_1$ as $\lambda_1(t_i^+)-\lambda_1(t_i^-) =
2(x_1(t_i)-y_i),$ at measurements $t_i$. The optimality conditions
on the gradients also result in the boundary conditions 
$\lambda_1(1)=0$ and $\lambda_2(1)=0$. Finally, the gradients with
respect to the initial conditions are
\begin{align}
\nabla_{x_1(0)}E  = -\lambda_1(0), \mbox{ and } \nabla_{x_2(0)}E = -\lambda_2(0)\enspace.
\label{eqn:euclidean_gradients_linefit}
\end{align}
These gradients are evaluated by integrating backward the adjoint
system \rk{to $t=0$ starting from $t=1$}.

This optimal-control perspective constitutes a 
general method for estimating first-order curves which 
allows to generalize
the notion of straight lines to manifolds (geodesics), as long as the
forward system (dynamics), the gradient computations, as well as the 
gradient steps all respect the geometry of the underlying space.

\subsection{Time-warped regression}
\label{subsection:time_warped_linear_regression}

\mn{Fitting straight lines is too restrictive for some data. Hence, the}
idea of time-warped regression is to use a simple model to warp
the time-points, or more generally the independent variable, when 
comparison to data is performed, \eg, as in the data
matching term of Eq. \eqref{eqn:euclidan_linefit_control}. The \textit{time-warp}
should maintain the order of the data, and hence needs to be
diffeomorphic. This is conceptually similar to an \textit{error-in-variables}
model where uncertainties in the independent variables are
modeled. However, in the concept of time-warping, we are not directly 
concerned with modeling such uncertainties, but instead in obtaining a somewhat
richer model based on a known and easy-to-estimate linear
regression model.  


In principle, the mapping of the time points could be described by a general diffeomorphism.
In fact, such an approach is followed in \cite{Durrleman13a} for
spatio-temporal atlas-building in the context of shape analysis. 
Our motivation for proposing an approach to linear regression with 
\textit{parametric}
time-warps is to keep the model simple while 
gaining more flexibility. Extensions to non-parametric
approaches can easily be obtained. A representative of a simple
parametric regression model is \textit{logistic regression}\footnote{Not to be
confused with the statistical classification method.} which is typically
used to model saturation effects. 
Under this model, points that are close in time for the linear fit may be mapped to points far apart in time, thereby allowing to model saturations for instance (\cf~Fig. \ref{fig:time_warped_least_square}). Other possibilities of
parametric time-warps include those derived from families of quadratic, 
logarithmic and exponential functions.

\rk{Formally, let $f:\mathbb{R} \rightarrow \mathbb{R}$, 
$t \mapsto \bar{t} = f(t; \boldsymbol{\theta})$ denote a 
parametrized (by $\boldsymbol{\theta}$) time-warping function} 
and let $x_1(\overline{t})$ 
denote the particle on the regression line in the warped time coordinates $\bar{t}$. 
Following this notation, the states are denoted as $(x_1(\overline{t}),
x_2(\overline{t}))$ and represent position and slope in re-parametrized
time $\overline{t}$. In \textit{time-warped regression}, the \textit{data matching} 
term of Eq. \eqref{eqn:euclidan_linefit_control} \rk{then} becomes
\begin{equation}
  \sum_{i=0}^{N-1}
  \|x_1(f(t_i;\boldsymbol{\theta}))-y_i\|^2
  \label{eq:tw_gr_energy_euc}\enspace,
\end{equation} 
and the objective (as before) is to optimize $x_1(\bar{t}_0)$
and $x_2(\bar{t}_0)$ as well as the parameter 
$\boldsymbol{\theta}$ of $f(t;\boldsymbol{\theta})$. 

A convenient way to minimize the energy functional in 
Eq. \eqref{eqn:euclidan_linefit_control} with the data matching term of 
Eq. \eqref{eq:tw_gr_energy_euc}, is to use an alternating optimization
strategy. That is,
we first fix $\boldsymbol{\theta}$ to update the initial conditions, and then fix the 
initial conditions to update $\boldsymbol{\theta}$.
\mn{This} requires the derivative of the energy with respect to $\boldsymbol{\theta}$
for fixed $x_1(\bar{t})$. Using the chain rule, we obtain the gradient 
$\nabla_{\boldsymbol{\theta}} E$
as
\begin{equation}
2\sum_{i=0}^{N-1}(x_1(f(t_i;\boldsymbol{\theta}))-y_i)^\top\dot{x}_1(f(t_i; \boldsymbol{\theta}))\nabla_{\boldsymbol{\theta}}f(t_i;\boldsymbol{\theta})\enspace.
\label{eq:tw_grad}
\end{equation}
Given a numerical
solution to the regression problem of Section~\ref{subsection:an_optimal_control_perspective}, the
time-warped extension alternatingly updates (a) the initial conditions $(x_1(\bar{t}_0),x_2(\bar{t}_0))$
in the warped time domain using the gradients in 
Eq. \eqref{eqn:euclidean_gradients_linefit} and (b) $\boldsymbol{\theta}$ using the
gradient in Eq.~\eqref{eq:tw_grad}.
Fig.~\ref{fig:time_warped_least_square} visualizes the principle of time-warped
linear regression on a collection of artificially generated data points. 
While the new model only slightly increases the overall complexity, it 
notably increases modeling flexibility by using a curve instead of a straight 
line.


\subsection{Cubic spline regression}
\label{subsection:cubic_spline_regression}

To further increase the flexibility of a regression model, cubic splines are another 
commonly used technique. In this section, we revisit cubic spline regression
in \rk{$\mathbb{R}^n$} from the optimal-control perspective.  
\mn{This will facilitate the transition to general Riemannian manifolds.}

\subsubsection{Variational formulation}
\label{subsubsection:variational_formulation}

An acceleration-controlled curve with time-dependent states 
$(x_1,x_2,x_3)$ such that \mn{$\dot{x}_1=x_2$ and $\dot{x}_2=x_3$}, 
defines a cubic curve in \rk{$\mathbb{R}^n$}. Such a curve is a solution
to the energy minimization problem, \cf~\cite{Ahlberg67a}, 
\begin{align}
\begin{split}
\min_{\boldsymbol{\Theta}} & \quad E(\boldsymbol{\Theta}) = 
\frac{1}{2}\int_0^1\|x_3\|^2~dt, \\
\text{subject to} & \quad  \mn{\dot{x}_1=x_2} \text{ and } \mn{\dot{x}_2=x_3}\enspace, \\
\end{split}
\label{eq:eucrelaxation}
\end{align}
with $\boldsymbol{\Theta} = \{x_i\mn{(t)}\}_{i=1}^3$. Here, $x_3$ is referred to as the \textit{control variable}
that describes the acceleration of the dynamics in this system. Similar to the strategy for 
fitting straight lines, we can get a relaxation solution to Eq.~\eqref{eq:eucrelaxation} by 
adding adjoint variables which leads to the system of adjoint 
equations $\dot{\lambda}_1 = 0$ and $\dot{x}_3 = -\lambda_1$. 

\subsubsection{From relaxation to shooting} 
\label{subsubsection:from_relaxation_to_shooting}

To obtain the shooting formulation, we explicitly add the evolution 
of $x_3$, \ie, $\dot{x}_3 = -\lambda_1$, as another 
dynamical constraint; this increases the order of the dynamics. 
Setting $x_4=-\lambda_1$ results in the classical system of equations for shooting cubic curves 
\begin{align}
\begin{aligned}
\dot{x}_1=x_2(t), & &  \dot{x}_2=x_3(t), & & \dot{x}_3=x_4(t), & & \dot{x}_4=0\enspace.
\end{aligned}\label{eq:shootingevolution}
\end{align}
The states $(x_1,x_2,x_3,x_4)$, at all times, are entirely determined
by their initial values $\{x_i(0)\}_{i=1}^4$ and, in particular we have
$x_1(t)=x_1(0) + x_2(0)t + \frac{1}{2}x_3(0)t^2+\frac{1}{6}x_4(0)t^3$.

\subsubsection{Data-independent controls}
\label{subsubsection:data_independent_controls}

Using the shooting equations of Eq.~\eqref{eq:shootingevolution} for 
cubic splines, we can define a
\emph{smooth} curve that best fits the data in the least-squares
sense. Since a cubic polynomial by itself is restricted to only fit
``cubic-like'' data, we add flexibility by gluing together piecewise 
cubic polynomials. Typically, we define controls at pre-defined
locations, and only allow the state $x_4$ to jump at those locations.

We let $\{t_c\}_{c=1}^C, t_c\in (0,1)$ denote $C$ data-independent 
fixed control points, which implicitly define $C+1$ intervals in 
$[0,1]$, denoted as $\{\mathcal{I}_c\}_{c=1}^{C+1}$. 
The constrained energy minimization problem corresponding
to the regression task, in this setting, can be written as
\begin{align}
\begin{split}
\min_{\boldsymbol{\Theta}} & \quad E(\boldsymbol{\Theta}) = \sum_{c=1}^{C+1}\sum_{i\in\mathcal{I}_c} \|x_1(t_i)-y_i\|^2, \\
\text{subject to} 	& \quad
	\begin{rcases}
  		&\dot{x}_1=x_2(t),~	 \dot{x}_2=x_3(t), \\
		&\dot{x}_3=x_4(t),~ \dot{x}_4=0, \\
	\end{rcases} \text{within $\mathcal{I}_c$} \\
	& \quad \mbox{and}~x_1, x_2, x_3 \text{ continuous across $t_c$}\enspace,
\end{split}
\label{eq:eucrelaxation2}
\end{align}
with parameters $\boldsymbol{\Theta} = \{ \{x_i(0)\}_{i=1}^4,\{x_4(t_c)\}_{c=1}^C \}$.
Using time-dependent adjoint states $\{\lambda_i\}_{i=1}^4$ for the dynamics constraints, 
and (time-independent) duals $\nu_{c,i}$ for the 
continuity constraints, \rk{we derive the adjoint system of 
equations from the unconstrained Lagrangian as}
\begin{equation}
\dot{\lambda}_1=0,~\dot{\lambda}_2=-\lambda_1,~\dot{\lambda}_3=-\lambda_2,~\dot{\lambda}_4=-\lambda_3\enspace.
\end{equation}

The gradients with respect to the initial conditions for states $\{x_i(0)\}_{i=1}^4$ are
\begin{align}
&\nabla_{x_1(0)}E = -\lambda_1(0), \quad \nabla_{x_2(0)}E  = -\lambda_2(0), \notag \\
&\nabla_{x_3(0)}E = -\lambda_3(0), \quad \nabla_{x_4(0)}E  = -\lambda_4(0)\enspace.
\end{align}
The \textit{jerks} (\ie, rate of acceleration change) at $x_4(t_c)$ are updated using $\nabla_{x_4(t_c)}E=-\lambda_4(t_c)$.  
The values of the adjoint variables at $0$ are computed by
integrating backward the adjoint system starting from $\lambda_i(1) =0$ for $i=1,\ldots,4$. Note that $\lambda_1$,
$\lambda_2$ and $\lambda_3$ are continuous at joints, but $\lambda_1$
jumps at the data-point location as per $\lambda_1(t^+_i)-\lambda_1(t^-_i) =
2(x_1(t_i)-y_i)$. During backward integration,
$\lambda_4$ starts with zero at each interval at $t_{c+1}$ and the
accumulated value at $t_c$ is used for the gradient update of
$x_4(t_c)$. 

It is critical to note that, along the time $t$, such a formulation
guarantees that: (a) $x_4(t)$ is piecewise constant, (b) $x_3(t)$ is
piecewise linear, (c) $x_2(t)$ is piecewise quadratic, and (d)
$x_1(t)$ is piecewise cubic. Thus, this results in a cubic spline
curve. Fig.~\ref{fig:cublic_spline_euclidean} demonstrates this
shooting-based spline fitting method on scalar-valued data. While 
it is difficult to explain this data with one simple cubic curve, it suffices to
add one control point to recover the underlying trend.
The state $x_4$ experiences a jump at the control location that
integrates up three-times to give a $C^2$-continuous evolution for the
state $x_1$.

\section{Regression on Riemannian Manifolds}
\label{section:geodesic_regression}

In this section, we adopt the optimal-control perspective of previous sections 
and generalize the regression problem\rk{s} to nonflat, smooth Riemannian manifolds.
\rk{In the literature this generalization is typically referred to as \textit{geodesic regression}.}
For a thorough treatment of Riemannian manifolds, we refer the reader to \cite{Boothby86a,
do1992riemannian}. We remark that the term geodesic regression \mn{here} does not refer to the model that is
fitted but rather to the fact that the Euclidean distance in the matching term of 
the energies is replaced by the geodesic distance on the manifold. In particular,
the measurements $\{y_i\}_{i=0}^{N-1}$ in Euclidean space now become elements 
$\{Y_i\}_{i=0}^{N-1}$ on some Riemannian manifold $\mathcal{M}$ with Riemannian
metric $\langle\cdot,\cdot\rangle_p$ at $p \in \mathcal{M}$\footnote{We omit the subscript $p$ when it is clear from the context.}. 
The geodesic distance, induced by this metric, 
will be denoted as $d_g$.
For generality, we also replace $t_i$ with $r_i$, indicating that the independent value 
does not have to be \textit{time}, but can also represent other entities, 
\eg, counts or speed. 

Our first objective is to estimate a geodesic $\gamma: \mathbb{R} \rightarrow \mathcal{M}$, 
represented by initial conditions $\gamma(r_0)$ (\ie, initial point), and $\dot{\gamma}(r_0)$
(\ie, initial velocity at the tangent space $\mathcal{T}_{\gamma(r_0)}\mathcal{M}$), while solving
\begin{align}
\begin{split}
\min_{\boldsymbol{\Theta}} E(\boldsymbol{\Theta}) = & \alpha \underbrace{\int_{0}^{1} \langle \dot{\gamma}, \dot{\gamma} \rangle_{\gamma(\mn{r})}~dr}_{\mbox{\small Regularity}} +
 \frac{1}{\sigma^2}\underbrace{\sum_{i=0}^{N-1} d_g^2(\gamma(r_i), Y_i)}_{\mbox{\small Data-matching }}\\
\mbox{subject to} & \quad \nabla_{\dot{\gamma}} \dot{\gamma} = 0 \mbox{ (geodesic equation)} \enspace,
\end{split}
\label{eqn:geodesic_regression_manifolds}
\end{align}
with $\boldsymbol{\Theta} = \{\gamma(0),\dot{\gamma}(0)\}$ and $\nabla$ denoting 
the Levi-Civita connection on $\mathcal{M}$. The 
covariant derivative $\nabla_{\dot{\gamma}}\dot{\gamma}$ of $0$ ensures that the curve
is a geodesic.
The parameters $\alpha \ge 0$ and $\sigma > 0$ balance the regularity and the data-matching
term. In the Euclidean case, there is typically no regularity term because we usually do not have prior knowledge about the slope. Similarly, on Riemannian manifolds we may penalize the initial velocity by choosing $\alpha>0$; but typically, $\alpha$ is also set to $0$. 
The regularity term on the velocity can be further reduced to a smoothness penalty at 
$r_0$, \ie, $\int_0^1 \langle \dot{\gamma}, \dot{\gamma} \rangle dr = \langle \dot{\gamma}(r_0), 
\dot{\gamma}(r_0) \rangle$, because of the energy conservation along the geodesic.
Also, since the geodesic is represented by the initial conditions
$(\gamma(r_0), \dot{\gamma}(r_0))$, we can move along the geodesic and estimate
the point $\gamma(r_i)$ that corresponds to $Y_i$.


\subsection{Optimization via geodesic shooting}
\label{sec:adjoint_optimization}


Taking the optimal-control point of view, the second-order problem of 
Eq.~\eqref{eqn:geodesic_regression_manifolds} can be \mn{written as a} system of first-order, 
upon the introduction of auxiliary states 
\begin{equation}
X_1(r) = \gamma(r), \quad \mbox{and} \quad X_2(r) = \dot{\gamma}(r)\enspace.\label{eqn:aux}
\end{equation}
Here, $X_1$ corresponds to the \textit{intercept} and $X_2$ corresponds to the 
\textit{slope} in classic linear regression. \rk{Considering the simplified 
smoothness penalty of the previous section, the original constrained minimization 
problem of Eq.~\eqref{eqn:geodesic_regression_manifolds} reduces to}
\begin{align}
  \begin{split}
  \min_{\boldsymbol{\Theta}} & \quad E(\boldsymbol{\Theta}) = 
  \alpha\langle X_2(r_0), X_2(r_0) \rangle~+ \\ 
  & \quad \frac{1}{\sigma^2}\sum_{i=0}^{N-1} d_g^2(X_1(r_i), Y_i)\\
  \mbox{subject to} & \quad \nabla_{X_2}X_2 =0\enspace,
  \end{split}
\label{eqn:geodesic_regression_unconstraint}
\end{align}
with $\boldsymbol{\Theta} = \{X_i(r_0)\}_{i=1}^2$. 
\rk{Note that} $X_1(r_i)$ is the estimated point on the geodesic at 
$r_i$, obtained by shooting forward with $X_1(r_0)$ and $X_2(r_0)$. 
Analogously to the elaborations of previous sections, we convert 
Eq.~\eqref{eqn:geodesic_regression_unconstraint} to 
an unconstrained minimization problem via time-dependent 
adjoint variables, then take variations with respect to its arguments 
and eventually get (1) dynamical systems of states and adjoint variables, 
(2) boundary conditions on the adjoint variables, and (3) gradients with respect to
initial conditions. By shooting forward~/~backward and \rk{updating 
the initial states via the gradients}, we can obtain a numerical solution
\rk{to the problem}.

\subsection{Time-warped regression}
\label{subsection:manifold_time_warping}

The time-warping strategy of Section~\ref{subsection:time_warped_linear_regression} 
can be also 
adapted to Riemannian manifolds, because it focuses on warping the axis of the 
independent scalar-valued 
variable, not the axis of the dependent manifold-valued variable.  
In other words, the time-warped model is independent 
of the underlying type of space. Formally, given a warping function $f$
(\cf~Section~\ref{subsection:time_warped_linear_regression}), all instances
of the form $X_i(r_i)$ in Eq. \eqref{eqn:geodesic_regression_unconstraint} 
are replaced by $X_i(f(r_i; \boldsymbol{\theta}))$.
While the model retains its simplicity, \ie, we still fit geodesic curves, 
the warping function allows for increased modeling flexibility.

Since we have an existing solution to the problem of fitting geodesic curves, 
the easiest way to minimize the resulting energy
is by alternating optimization, similar to 
Section~\ref{subsection:time_warped_linear_regression}. 
This requires the derivative of the energy 
with respect to $\boldsymbol{\theta}$ for fixed $X(\overline{r})$. While \rk{the derivation}
is slightly more involved, application of the chain rule and \cite[Appendix A]{Samir12a} yields
\begin{align}
\begin{split}
& \nabla_{\boldsymbol{\theta}} E = 2\alpha \langle \dot{X}_2(f(r_0; \theta)), X_2(f(r_0; \theta)) \rangle \nabla_{\boldsymbol{\theta}} f(r_0;\boldsymbol{\theta}) \\
& -\frac{2}{\sigma^2}\sum_{i=0}^{N-1}\langle \rlog_{X_1(f(r_i;\boldsymbol{\theta}))}Y_i, \dot{X}_1(f(r_i;\boldsymbol{\theta})) \rangle \nabla_{\boldsymbol{\theta}} f(r_i;\boldsymbol{\theta})
\end{split} 
\label{eqn:energy_theta_gradient_manifold}
\end{align}
where $\rlog_{X_1(f(r_i;\boldsymbol{\theta}))}Y_i$ denotes the Riemannian log-map, 
\ie, the initial velocity of the geodesic connecting $X_1(f(r_i;\boldsymbol{\theta}))$ 
and $Y_i$ in unit time and $\dot{X}_1(f(r_i;\boldsymbol{\theta}))$ 
is the velocity of the regression geodesic at the warped-time point. 
This leaves to choose a good parametric model for 
$f(r;\boldsymbol{\theta})$. As we require the time warp
to be diffeomorphic, we choose a 
parametric model which is diffeomorphic by construction. One possible choice is the 
generalized logistic function~\cite{Fekedulegn99a}, \eg, with asymptotes 0 for $r\rightarrow -\infty$ and 
1 for $r\rightarrow \infty$, given by
\begin{equation}
  f(r; \boldsymbol{\theta}) = \frac{1}{(1+\beta e^{-k(r-M)})^{1/m}}\enspace ,
  \label{eqn:generalized_logistic}
\end{equation}
with $\boldsymbol{\theta} = (k, M, \beta, m)$. The parameter $k$ controls the growth rate, 
$M$ is the time of maximum growth if $\beta = m$, $\beta$ and $m$ define the value of $f$ 
at $t=M$, and $m>0$ affects the asymptote of maximum growth. By using this function, 
we map the original infinite time interval to a warped time-range from $0$ to $1$. 
In summary, the algorithm using alternating optimization is as follows:
\begin{itemize}
  \item[0)] Initialize $\boldsymbol{\theta}$ such that the warped time is evenly distributed within (0, 1). 
  \item[1)] Compute 
  $\{\overline{r}_i = f(r_i;\boldsymbol{\theta})\}_{i=0}^{N-1}$
  and perform standard geodesic regression using the new time-points.
  \item[2)] Update $\boldsymbol{\theta}$ by numerical optimization using the gradient given in Eq.~\eqref{eqn:energy_theta_gradient_manifold}. 
  \item[3)] Check convergence. If not converged goto 1).
\end{itemize}

\subsection{Cubic spline regression}
\label{subsection:geodesic_regression_higher_order_models}

Similar to Section~\ref{subsection:cubic_spline_regression}, 
cubic curves on a Riemannian manifold $\mathcal{M}$ 
can be defined as solutions
to the variational problem of minimizing an accelleration-based
energy.
The notion of acceleration is defined using the
covariant derivatives on Riemannian
manifolds~\cite{noakes1989cubic,camarinha1995splines}. 
In particular, we define a \textit{time-dependent control}, \rk{\ie}, a 
forcing variable $X_3(\mn{r})$, as 
\begin{align*} 
X_3(\mn{r})=\nabla_{X_2(\mn{r})}X_2(\mn{r}) =\nabla_{\dot{X}_1(\mn{r})}\dot{X}_1(\mn{r})
\enspace.
\end{align*} 
We can interpret $X_3(\mn{r})$ as a
control that forces the curve $X_1(\mn{r})$ to deviate from being a
geodesic~\cite{trouve2012splines} (which is 
the case if $X_3(r)=0$). As an end-point problem, a
Riemannian \rk{cubic curve} is thus defined by the curve $X_1(\mn{r})$ such 
that it minimizes an energy of the form
\begin{align*}
 E(X_1) = \frac{1}{2}\int^1_0
 \|\nabla_{\dot{X}_1}\dot{X}_1\|^2 dt,
\end{align*}
where the norm $\| \cdot \|$ is induced by the metric on $\mathcal{M}$ at
$X_1$.
In Section~\ref{subsection:grassmann_cubic_spline_regression},
this concept will be adapted to the Grassmannian to enable regression
with cubic splines.


\begin{algorithm*}[t!]
\KwData{$\{(r_i,\mathbf{Y}_i)\}_{i=0}^{N-1}$, $\alpha$ and $\sigma^2$}
\KwResult{$\mathbf{X}_1(r_0)$, $\mathbf{X}_2(r_0)$}
Set initial $\mathbf{X}_1(r_0)$ and $\mathbf{X}_2(r_0)$, \eg, $\mathbf{X}_1(r_0) = \mathbf{Y}_0$, and $\mathbf{X}_2(r_0) = 0$. \\ 
\While{not converged}{
  Solve Eqs.~\eqref{eqn:grassmannian_dynamics} with $\mathbf{X}_1(r_0)$ and $\mathbf{X}_2(r_0)$
  forward for $r\in[r_0,r_{N-1}]$. \\
  Solve
  $\begin{cases}
    \dot{\lambda}_1 = \lambda_2 \mathbf{X}_2^\top\mathbf{X}_2,~\lambda_1(r_{N-1}+) = 0,\\
    \dot{\lambda}_2 = -\lambda_1 + \mathbf{X}_2(\lambda_2^\top\mathbf{X}_1 + \mathbf{X}_1^\top\lambda_2),~\lambda_2(r_{N-1}) = 0
  \end{cases}$
  backward with jump conditions
    $\lambda_1(r_i-) = \lambda_1(r_i+) -\frac{1}{\sigma^2}\nabla_{\mathbf{X}_1(r_i)}d_g^2(\mathbf{X}_1(r_i),\mathbf{Y}_i)$,  
and $\nabla_{\mathbf{X}_1(r_i)}d_g^2(\mathbf{X}_1(r_i),\mathbf{Y}_i)$ computed as $-2\rlog_{\mathbf{X}_1(r_i)}(\mathbf{Y}_i)$.
For multiple measurements at a given $r_i$, the jump conditions for each measurement are added up.\\
  Compute gradients with respect to initial conditions:
  \begin{eqnarray}
    \nabla_{\mathbf{X}_1(r_0)}E &=& -(\mathbf{I}_n-\mathbf{X}_1(r_0)\mathbf{X}_1(r_0)^\top)\lambda_1(r_0-) + \mathbf{X}_2(r_0)\lambda_2(r_0)^\top\mathbf{X}_1(r_0),  \notag \\ 
    \nabla_{\mathbf{X}_2(r_0)}E &=& 2\alpha \mathbf{X}_2(r_0) - (\mathbf{I}_n-\mathbf{X}_1(r_0)\mathbf{X}_1(r_0)^\top)\lambda_2(r_0). \notag
  \end{eqnarray}
  Use a line search with these gradients to update $\mathbf{X}_1(r_0)$ and $\mathbf{X}_2(r_0)$ (see \mn{supplementary} material).
}
\caption{Standard Grassmannian geodesic regression (Std-GGR)}
\label{alg:geodesic_regression}
\end{algorithm*}

\section{Regression on the Grassmannian}
\label{section:grassmannian_geodesic_regression}

The Grassmannian is a type of Riemannian manifold where 
the geodesic distance, parallel transport, as well as the Riemannian log-/exp-map
are relatively simple to compute \cite{Gallivan03a}. Before specializing our three 
regression models to this manifold, we first discuss its Riemannian structure
in Section \ref{subsection:riemannian_structure_of_the_grassmannian} (see 
\cite{Absil04a} for a thorough treatment) and 
review how different types of data can be represented on the Grassmannian
in Section \ref{subsection:representation_on_the_grassmannian}. 

\subsection{Riemannian structure of the Grassmannian}
\label{subsection:riemannian_structure_of_the_grassmannian}

The \textit{Grassmann} manifold $\mathcal{G}(p,n)$ is defined as the set of $p$-dimensional 
linear subspaces of $\mathbb{R}^n$, typically represented by an orthonormal matrix 
$\mathbf{Y} \in \mathbb{R}^{n \times p}$, such that the column vectors span 
$\mathcal{Y}$, \ie, $\mathcal{Y} = \myspan(\mathbf{Y})$. 
It can equivalently be defined as a quotient space within the special 
orthogonal group $SO(n)$ as $\mathcal{G}(p,n) := \mathcal{SO}(n)/(\mathcal{SO}(n-p) 
\times \mathcal{SO}(p))$. The \textit{canonical metric}  
$g_\mathcal{Y}: \mathcal{T}_\mathcal{Y}\mathcal{G}(p,n) \times \mathcal{T}_\mathcal{Y}\mathcal{G}(p,n) 
\rightarrow \mathbb{R}$ on $\mathcal{G}(p,n)$ is given by
\begin{equation}
g_\mathcal{Y}(\mathbf{\Delta}_\mathcal{Y},\mathbf{\Delta}_\mathcal{Y}) =  \tr \mathbf{\Delta}_\mathcal{Y}^\top \mathbf{\Delta}_\mathcal{Y} = \tr
\mathbf{C}^\top(\mathbf{I}_n -\mathbf{Y}\mathbf{Y}^\top)\mathbf{C}\enspace,
\label{eqn:grmetric}
\end{equation}
where $\mathbf{I}_n$ denotes the $n \times n$ identity matrix, 
$\mathcal{T}_\mathcal{Y}\mathcal{G}(p,n)$ is the tangent space at $\mathcal{Y}$,
$\mathbf{C} \in \mathbb{R}^{n \times p}$ is arbitrary, and $\mathbf{Y}$
is a \textit{representer} for $\mathcal{Y}$.
Under this choice of metric, the arc-length of the geodesic connecting two 
subspaces $\mathcal{Y},\mathcal{Z} \in \mathcal{G}(p,n)$ is related to the 
\textit{canonical angles} $\phi_1,\ldots\phi_p \in [0,\pi/2]$
between $\mathcal{Y}$ and $\mathcal{Z}$ as
$d_g^2(\mathcal{Y},\mathcal{Z}) = ||\boldsymbol{\phi}||_2^2$.
In what follows, we slightly change notation and use $d_g^2(\mathbf{Y},
\mathbf{Z})$, with $\mathcal{Y}=\myspan(\mathbf{Y})$ and $\mathcal{Z}=\myspan(\mathbf{Z})$.
In fact, the (squared) geodesic distance can be computed from the SVD
decomposition $\mathbf{U}(\cos \boldsymbol{\Sigma})\mathbf{V}^\top = \mathbf{Y}^\top\mathbf{Z}$ as
$d_g^2(\mathbf{Y}, \mathbf{Z}) = ||\cos^{-1}(\mydiag \boldsymbol{\Sigma})||^2$ (cf. 
\cite{Gallivan03a}), where $\mathbf{\Sigma}$ is a diagonal matrix with principal
angles $\phi_i$.

Finally, consider a curve $\gamma: [0,1] \rightarrow \mathcal{G}(p,n), r \mapsto \gamma(r)$ such that $\gamma(0) = \mathcal{Y}_0$ and $\gamma(1) = \mathcal{Y}_1$, with $\mathcal{Y}_0$
represented by $\mathbf{Y}_0$ and $\mathcal{Y}_1$ represented by $\mathbf{Y}_1$. 
The \textit{geodesic equation} for such a curve, given that 
$\dot{\mathbf{Y}} =\nicefrac{d}{d\mn{r}}\mathbf{Y}(\mn{r}) \doteq
 (\mathbf{I}_n -\mathbf{Y}\mathbf{Y}^\top)\mathbf{C}$, on $\mathcal{G}(p,n)$ is given by
\begin{equation}
\begin{split}
\ddot{\mathbf{Y}}(r) + \mathbf{Y}(r)[\dot{\mathbf{Y}}(r)^\top\dot{\mathbf{Y}}(r)]= 0\enspace,
\end{split}
\label{eq:geodesic_equation}
\end{equation}
which also defines the Riemannian exponential map on the Grassmannian  as an ODE for convenient 
numerical computations. Integrating Eq.~\eqref{eq:geodesic_equation}, starting 
with initial conditions, ``shoots'' the geodesic forward in time.

\subsection{Representation on the Grassmannian}
\label{subsection:representation_on_the_grassmannian}

We particularly describe two types of data objects that can be represented
as points on the Grassmannian: linear dynamical systems (LDS) and shapes.


\subsubsection{Linear dynamical systems}
\label{sec:lds}
In the computer vision literature, \textit{dynamic texture} models \cite{dorettoCWS03} are commonly 
applied to model videos as realizations of linear dynamical systems (LDS). For a video, represented by a collection of vectorized frames $\mathbf{y}_1,\ldots,\mathbf{y}_\tau$ with $\mathbf{y}_i \in \mathbb{R}^n$,
the standard dynamic texture model with $p$ states has the form
\begin{align}
  \mathbf{x}_{k+1} 	= & ~\mathbf{A} \mathbf{x}_k + \mathbf{w}_k, \quad  	& 	\mathbf{w}_k\sim \mathcal{N}(0,\mathbf{W}), \notag \\
  \mathbf{y}_{k}		= & ~\mathbf{C} \mathbf{x}_k + \mathbf{v}_k, \quad  	&	\mathbf{v}_k\sim \mathcal{N}(0,\mathbf{R}) \enspace,
\end{align}
with $\mathbf{x}_k \in \mathbb{R}^p, \mathbf{A} \in \mathbb{R}^{p \times p}$, and $\mathbf{C} \in \mathbb{R}^{n \times p}$. When relying on the prevalent estimation approach of \cite{dorettoCWS03}, the matrix $\mathbf{C}$ is, by design, of (full) rank $p$ (\ie, the number of states) and by construction we obtain an \textit{observable} system, where a full rank \textit{observability} matrix 
$\mathbf{O} \in \mathbb{R}^{np \times p}$ is defined as
$\mathbf{O} = [\mathbf{C}~(\mathbf{C}\mathbf{A})~(\mathbf{C}\mathbf{A}^2)~\cdots~(\mathbf{C}\mathbf{A}^{p-1})]^\top$.
This system identification is not unique because systems $(\mathbf{A},\mathbf{C})$ and $(\mathbf{T}\mathbf{A}\mathbf{T}^{-1}, \mathbf{C}\mathbf{T}^{-1})$ with $\mathbf{T} \in \mathcal{GL}(p)$\footnote{$\mathcal{GL}(p)$ is the general linear group of $p\times p$ invertible matrices.} have the same transfer function. Hence, the realization subspace spanned by $\mathbf{O}$ is a point on the Grassmannian $\mathcal{G}(p,n)$ and the observability matrix is a representer of this subspace. We identify an LDS model for a video by its $np \times p$ orthonormalized observability matrix.

\subsubsection{Shapes}
\label{subsubsection:shapes}
We consider shapes as represented by a collection of $m$ landmarks. 
A \textit{shape matrix} is constructed from its $m$ landmarks
as $\mathbf{L} = [ (x_1, y_1, ...); (x_2, y_2, ...);\ldots; (x_m, y_m, ...) ]$. 
Using SVD on this matrix, \ie, $\mathbf{L}=\mathbf{U}\mathbf{\Sigma}\mathbf{V}^\top$, 
we obtain an affine-invariant shape representation from the left-singular vectors 
$\mathbf{U}$~\cite{Begelfor06a,Sepiashvili03a}. This establishes a mapping from 
the shape matrix to a point on the Grassmannian (with $\mathbf{U}$ as the 
representative). Such a representation has been used for facial aging 
regression for instance \cite{Turuga10a}.

\begin{algorithm*}[t!]
\KwData{$\{(r_i,\mathbf{Y}_i)\}_{i=0}^{N-1}$, $\{r_c\}_{c=1}^C$, $\alpha$ and $\sigma^2$}
\KwResult{$\mathbf{X}_1(r_0)$, $\mathbf{X}_2(r_0)$, $\mathbf{X}_3(r_0)$, $\mathbf{X}_4(r_0)$, $\{\mathbf{X}_4(r_c^+)\}_{c=1}^C$}
Set initial $\mathbf{X}_1(r_0)$ as $\mathbf{Y}_0$ for example,  and $\mathbf{X}_2(r_0)$, $\mathbf{X}_3(r_0)$, $\mathbf{X}_4(r_0)$, $\{\mathbf{X}_4(r_c^+)\}_{c=1}^C$ as zero matrices.\\
\While{not converged}{
  Solve Eq.~\eqref{eqn:shooting_cubic_curve_grassmannian}
  forward in each interval with $\mathbf{X}_1(r_0)$, $\mathbf{X}_2(r_0)$, $\mathbf{X}_3(r_0)$, $\mathbf{X}_4(r_0)$, $\{\mathbf{X}_4(r_c^+)\}_{c=1}^C$, and $\{\mathbf{X}_1(r_c^+) = \mathbf{X}_1(r_c^-)$, $\mathbf{X}_2(r_c^+) = \mathbf{X}_2(r_c^-)$, $\mathbf{X}_3(r_c^+) = \mathbf{X}_3(r_c^-)\}_{c=1}^C$. \\
  Solve
  $\begin{cases}
    \dot{\lambda_1} &= \lambda_2 \mathbf{X}_2^\top \mathbf{X}_2 - \lambda_3 (\mathbf{X}_4^\top \mathbf{X}_1 - \mathbf{X}_3^\top \mathbf{X}_2) - \mathbf{X}_4 (\lambda_3^\top \mathbf{X}_1  + \mathbf{X}_2^\top \lambda_4), \\
    \dot{\lambda_2} &= -\lambda_1 + \mathbf{X}_2 (\lambda_2^\top \mathbf{X}_1 + \mathbf{X}_1^\top \lambda_2 -  \lambda_4^\top \mathbf{X}_3 - \mathbf{X}_3^\top \lambda_4) + \mathbf{X}_3 (\lambda_3^\top \mathbf{X}_1  + \mathbf{X}_2^\top \lambda_4) + \lambda_4 (-\mathbf{X}_1^\top \mathbf{X}_4 + \mathbf{X}_2^\top \mathbf{X}_3), \\
    \dot{\lambda_3} &= -\lambda_2 - \lambda_4 \mathbf{X}_2^\top \mathbf{X}_2  + \mathbf{X}_2 (\mathbf{X}_1^\top \lambda_3 + \lambda_4^\top \mathbf{X}_2) + 2 \alpha \mathbf{X}_3,  \\
    \dot{\lambda_4} &= \lambda_3 - \mathbf{X}_1 (\mathbf{X}_1^\top \lambda_3 + \lambda_4^\top \mathbf{X}_2)
  \end{cases}$
  backward with $\lambda_1(r_{N-1}) = \lambda_2(r_{N-1}) = \lambda_3(r_{N-1}) = \lambda_4(r_{N-1}) = \lambda_4(r_c^-) = 0$, and\\
 $\{\lambda_1(r_c^-) = \lambda_1(r_c^+)$, $\lambda_2(r_c^-) = \lambda_2(r_c^+)$, $\lambda_3(r_c^-) = \lambda_3(r_c^+)\}_{c=1}^C$, as well as jump conditions
    $\lambda_1(r_i^-) = \lambda_1(r_i^+) -\frac{1}{\sigma^2}\nabla_{\mathbf{X}_1(r_i)}d_g^2(\mathbf{X}_1(r_i),\mathbf{Y}_i)$,  
and $\nabla_{\mathbf{X}_1(r_i)}d_g^2(\mathbf{X}_1(r_i),\mathbf{Y}_i)$ computed as $-2\rlog_{\mathbf{X}_1(r_i)}(\mathbf{Y}_i)$. 
For multiple measurements at a given $r_i$, the jump conditions for each measurement are added up.\\
  Compute gradients with respect to initial conditions and the fourth state at control points:
  \begin{eqnarray}
    \nabla_{\mathbf{X}_1(r_0)}E &=& -(\mathbf{I}_n-\mathbf{X}_1(r_0) \mathbf{X}_1(r_0)^\top) \lambda_1(r_0^-) + \mathbf{X}_2(r_0) \lambda_2(r_0)^\top \mathbf{X}_1(r_0) + \mathbf{X}_3(r_0) \lambda_3(r_0)^\top \mathbf{X}_1(r_0), \notag\\
    \nabla_{\mathbf{X}_2(r_0)}E &=& -(\mathbf{I}_n - \mathbf{X}_1(r_0) \mathbf{X}_1(r_0)^\top) \lambda_2(r_0), \notag \quad \quad
    \nabla_{\mathbf{X}_3(r_0)}E = -(\mathbf{I}_n - \mathbf{X}_1(r_0) \mathbf{X}_1(r_0)^\top) \lambda_3(r_0), \notag \\
    \nabla_{\mathbf{X}_4(r_0)}E &=& -\lambda_4(r_0), \notag \quad \quad \quad \quad \quad \quad \quad \quad \quad \quad
    \quad \; \,   \nabla_{\mathbf{X}_4(r_c^+)}E = -\lambda_4(r_c^+),~c = 1...C \enspace. \notag
  \end{eqnarray}
  Use a line search with these gradients to update $\mathbf{X}_1(r_0)$, $\mathbf{X}_2(r_0)$, $\mathbf{X}_3(r_0)$, $\mathbf{X}_4(r_0)$, and $\{\mathbf{X}_4(r_c^+)\}_{c=1}^C$.
}
\caption{Cubic-spline Grassmannian geodesic regression (CS-GGR)}
\label{alg:spline_regression}
\end{algorithm*}

\subsection{Standard geodesic regression}
\label{subsection:grassmann_geodesic_regression}
 
We start by adapting the generic inner-product and the squared geodesic distance 
in Eq.~\eqref{eqn:geodesic_regression_manifolds} to the Riemannian structure of
$G(p,n)$. Given the auxiliary states of Eq. \eqref{eqn:aux},
now denoted as matrices $\mathbf{X}_1$ (initial point) and $\mathbf{X}_2$ (velocity), 
we can write the geodesic equation of Eq.~\eqref{eq:geodesic_equation} as a system 
of first-order dynamics:
\begin{align}
	\begin{split}
	\dot{\mathbf{X}}_1 = \mathbf{X}_2, \quad \mbox{and} \quad
	\dot{\mathbf{X}}_2 = -\mathbf{X}_1 (\mathbf{X}_2^\top  \mathbf{X}_2) \enspace.
	\end{split}
\label{eqn:grassmannian_dynamics}
\end{align}
For a point on the Grassmannian, it should further hold that (1)
$\mathbf{X}_1(r)^\top \mathbf{X}_1(r) = \mathbf{I}_p$ and (2) 
the velocity at $\mathbf{X}_1(r)$ needs to be orthogonal to that point, 
\ie, $\mathbf{X}_1(r)^\top \mathbf{X}_2(r) = \mathbf{0}$. 
If we enforce these two constraints at the starting point $r_0$, they 
will remain satisfied along the geodesic. Hence, the constrained 
minimization problem for standard Grassmannian geodesic regression is
\begin{align}
  \begin{split}
  \min_{\boldsymbol{\Theta}} & \quad  E(\boldsymbol{\Theta}) =
  \alpha~\tr \mathbf{X}_2(r_0)^\top \mathbf{X}_2(r_0)~+ \\
  & \quad 
    \frac{1}{\sigma^2} \sum_{i=0}^{N-1} d_g^2(\mathbf{X}_1(r_i), \mathbf{Y}_i)\\
  \mbox{subject to} & \quad \mathbf{X}_1(r_0)^\top \mathbf{X}_1(r_0) = \mathbf{I}_p, \\
  & \quad \mathbf{X}_1(r_0)^\top \mathbf{X}_2(r_0) = \mathbf{0} \mbox{ and Eq. \eqref{eqn:grassmannian_dynamics}} \enspace,
	\end{split}
\label{eqn:geodesic_regression_grassmannian}
\end{align}
with $\boldsymbol{\Theta} = \{ \mathbf{X}_i(r_0) \}_{i=1}^2$.
As \mn{in} previous sections, based on the adjoint method we obtain the shooting solution to Eq.~\eqref{eqn:geodesic_regression_grassmannian}, listed in Algorithm~\ref{alg:geodesic_regression}. 
We notice that the jump conditions on $\lambda_1$ involve the gradient of the residual term
$d^2_g(\mathbf{X}_1(r_i),\mathbf{Y}_i)$ with respect to $\mathbf{X}_1(r_i)$, \ie, the base
point of the residual on the fitted geodesic (see \mn{supplementary} material,
the gradient is $-2\rlog_{\mathbf{X}_1(r_i)}(\mathbf{Y}_i)$).
We refer to this problem of fitting a geodesic curve as \textit{standard Grassmannian
geodesic regression (Std-GGR)}.

\subsection{Time-warped regression}
\label{subsection:grassmann_time_warping}

Since the concept of time-warped geodesic regression is general for 
Riemannian manifolds, specialization to the Grassmannian 
is straightforward. We only need to use the Std-GGR solution
during the alternating optimization steps. By choosing the generalized 
logistic function of Eq. \eqref{eqn:generalized_logistic} 
to account for saturations of scalar-valued outputs, the time-warped 
model on $G(p,n)$ can be used to capture saturation effects \mn{for} which standard geodesic regression is insensible. We refer to this strategy as \textit{time-warped Grassmannian geodesic
regression (TW-GGR)}.

\renewcommand{\arraystretch}{1.1}
\begin{table*}
\centering
\begin{tabular}{|r|cccccc|ccc|}
\hline
\textbf{Method}         &  $\mathbf{X}_1(r_0)$ & $\mathbf{X}_2(r_0)$ & $\mathbf{X}_3(r_0)$ & $\mathbf{X}_4$s & $k$ & $M$ & MSD (G.T. \textit{vs.} Data) & MSD (Est. \textit{vs.} Data) & MSD (G.T. \textit{vs.} Est.)\\
\hline
Std-GGR  	& 0.0207  & 0.1124 & -- & --  & -- & -- & 7.0e-3 & 6.6e-3 & 0.3e-3 \\
TW-GGR 	&  0.0206 & 0.1619 & -- & -- & 0.0524 & 0.0056 & 6.9e-3 & 6.6e-3 & 0.3e-3\\
CS-GGR 	& 0.0672 & 0.5389 & 0.3600 & 0.9687 & -- & -- & 6.8e-3 & 5.8e-3 & 1.1e-3 \\
\hline
\end{tabular}
\caption{\small{Regression comparison with respect to (1) the initial conditions, 
(2) the parameters of the time-warp function ($k,M$),
and (3) mean square distance (MSD) among the ground truth (G.T.), estimated regression curves (Est.), and
data points. For $\mathbf{X}_1$: geodesic distance on the Grassmannian; for $\mathbf{X}_2$, 
$\mathbf{X}_3$, $\mathbf{X}_4$: $\|\mathbf{X}_i^{Est.} - \mathbf{X}_i^{G.T.}\|_F / \|\mathbf{X}_i^{G.T.}\|_F$; 
for multiple $\mathbf{X}_4s$, the mean is reported.\label{tab:synthetic_data_test}}}
\end{table*}
 
\subsection{Cubic spline regression}
\label{subsection:grassmann_cubic_spline_regression}

To enable cubic spline regression on the Grassmannian, we follow Section \ref{subsection:geodesic_regression_higher_order_models} and add
the external force $\mathbf{X}_3$. In other words, we represent an acceleration-controlled 
curve $\mathbf{X}_1(r)$ on \rk{$G(p,n)$} using a dynamic system with states 
$(\mathbf{X}_1, \mathbf{X}_2, \mathbf{X}_3)$ such that
\begin{align}
 \mathbf{X}_2 = \dot{\mathbf{X}}_1, \quad \mbox{and} \quad
 \mathbf{X}_3 = \dot{\mathbf{X}}_2 + \mathbf{X}_1(\mathbf{X}_2^\top\mathbf{X}_2)\enspace. \label{eqn:grassmann_accel}
\end{align}
Note that if $\mathbf{X}_3 = \mathbf{0}$, the second equation is 
reduced to the geodesic equation of Eq.~\eqref{eq:geodesic_equation}; this indicates that the curve is acceleration-free. To obtain an acceleration-controlled
curve, we need to solve the following constrained minimization problem:
\begin{align}
\begin{split}
\min_{\boldsymbol{\Theta}} 	& \quad E(\boldsymbol{\Theta}) = \frac{1}{2} \int_0^1 \tr \mathbf{X}_3^\top \mathbf{X}_3~dr \\
\text{subject to} 		        
					           &\quad \mathbf{X}_1^\top \mathbf{X}_1 = \mathbf{I}_p, 
						   \, \mathbf{X}_1^\top \mathbf{X}_2 = \mathbf{0}, 
						   \, \mbox{and Eq.~\eqref{eqn:grassmann_accel}}
\end{split}
\label{eqn:spline_curve_on_the_grassmannian}
\end{align}
with $\boldsymbol{\Theta} = \{\mathbf{X}_i\rk{(r_0)}\}_{i=1}^3$. The relaxation solution
to this problem gives us (see \mn{supplementary} material)
the system of 
equations for shooting cubic curves on 
$G(p,n)$:
\begin{align}
\begin{split}
\dot{\mathbf{X}}_1 &= \mathbf{X}_2,  \quad
\dot{\mathbf{X}}_2 = \mathbf{X}_3 - \mathbf{X}_1 \mathbf{X}_2^\top \mathbf{X}_2, \\
\dot{\mathbf{X}}_3 &= -\mathbf{X}_4 + \mathbf{X}_1 \mathbf{X}_1^\top \mathbf{X}_4 - \mathbf{X}_1 \mathbf{X}_2^\top \mathbf{X}_3, \\
\dot{\mathbf{X}}_4 &= \mathbf{X}_3 \mathbf{X}_2^\top \mathbf{X}_2 + \mathbf{X}_2 \mathbf{X}_4^\top \mathbf{X}_1 - \mathbf{X}_2 \mathbf{X}_3^\top \mathbf{X}_2\enspace.
\label{eqn:shooting_cubic_curve_grassmannian}
\end{split}
\end{align}

It is important to note that $\mathbf{X}_1$ does not follow a geodesic path under 
non-zero force $\mathbf{X}_3$. Hence, the constraints 
$\mathbf{X}_1(r)^\top \mathbf{X}_1(r) = \mathbf{I}_p$ and 
$\mathbf{X}_1(r)^\top \mathbf{X}_2(r) = \mathbf{0}$ should be 
enforced at \emph{every} instance of $r$ to keep the path on the
manifold. However, we can show (see \mn{supplementary} material) that 
enforcing $\mathbf{X}_1(r)^\top \mathbf{X}_2(r) = \mathbf{0}$ 
at all times already guarantees that $\mathbf{X}_1(r)^\top \mathbf{X}_1(r) = \mathbf{I}_p$ 
if this holds initially at $r=0$. Also, $\mathbf{X}_1(r)^\top \mathbf{X}_2(r) = \mathbf{0}$ implies that
$\mathbf{X}_1(r)^\top \mathbf{X}_3(r) = \mathbf{0}$. By using this fact during relaxation, the constraints
are already implicitly captured in Eqs.~\eqref{eqn:shooting_cubic_curve_grassmannian}.
Subsequently, for shooting we only need to guarantee that all these constraints
hold initially.

To get a cubic spline curve, we follow Section \ref{subsubsection:data_independent_controls} 
and introduce control points $\{r_c\}_{c=1}^C$, which divide the support of the 
independent variable into several intervals $\mathcal{I}_c$. The first three states
should be continuous at the control points, 
but the state $\mathbf{X}_4$ is allowed to jump. Hence, the spline regression 
problem on $G(p,n)$ becomes, \cf~Eq.\eqref{eq:eucrelaxation2}, 
\begin{align}
\begin{split}
\min_{\boldsymbol{\Theta}} & \quad  E(\boldsymbol{\Theta}) = 
	\alpha~\int_{r_0}^{r_{N-1}}\tr \mathbf{X}_3^\top \mathbf{X}_3~dr~+ \\
	& \quad \frac{1}{\sigma^2} \sum_{i=0}^{N-1} d_g^2(\mathbf{X}_1(r_i), \mathbf{Y}_i)\\
\text{subject to}&	\quad \mathbf{X}_1(r_0)^\top \mathbf{X}_1(r_0) = \mathbf{I}_p, \\			   & \quad \mathbf{X}_1(r_0)^\top \mathbf{X}_2(r_0) = \mathbf{0}, \\
			   &\quad \mathbf{X}_1(r_0)^\top \mathbf{X}_3(r_0) = \mathbf{0}, \\
			& \quad \mathbf{X}_1, \mathbf{X}_2, \mathbf{X}_3~\mbox{are continuous at $\{r_c\}_{c=1}^C$}, \\
					& \quad \mbox{and Eqs.~\eqref{eqn:shooting_cubic_curve_grassmannian} holds in each $\mathcal{I}_c$} \enspace,
\label{eqn:grassmann_spline_constrained}
\end{split}
\end{align}
with $\boldsymbol{\Theta} = \{ \{\mathbf{X}_i(r_0)\}_{i=1}^{4}, \{\mathbf{X}_4(r_c^+)\}_{c=1}^C \}$.
Algorithm~\ref{alg:spline_regression} lists the shooting 
solution to Eq.~\eqref{eqn:grassmann_spline_constrained}, referred to as 
\textit{cubic-spline Grassmannian geodesic regression (CS-GGR)}.

\section{Experiments on toy-examples}
\label{section:experiments_on_toy_examples}

We first demonstrate 
Std-GGR, TW-GGR and CS-GGR on synthetic data. 
Each data point represents a 
2D sine/cosine signal, sampled at $630$ points in $[0,10\pi]$ and embedded 
in $\mathbb{R}^{24}$. \rk{In particular}, the 2D signal $\mathbf{s} \in \mathbb{R}^{2 \times 630}$ 
is linearly projected via $\overline{\mathbf{s}} = \mathbf{U}\mathbf{s}$,
where $\mathbf{W} \sim \mathcal{N}(\mathbf{0},\mathbf{I}_{24})$ and 
$\mathbf{W} = \mathbf{U}\mathbf{\Sigma}\mathbf{V}^\top$. Finally, white 
Gaussian noise with $\sigma=0.1$ is added to $\overline{\mathbf{s}}$. For each
signal $\overline{\mathbf{s}} \in \mathbb{R}^{24 \times 630}$, we estimate
a (two-state, $p=2$) LDS as discussed in Section~\ref{sec:lds}, and use the corresponding observability matrix to represent it as a point on $G(2, 48)$. 
Besides, each data point has an associated scalar value; this independent 
variable is uniformly distributed within $(0, 10)$ and controls the signal 
frequency of the data point. For Std-GGR, we directly use \rk{this} value 
as the signal frequency to generate 2D sine/cosine signals, while for TW-GGR and 
CS-GGR, a generalized logistic function or a sine function is adopted to convert 
the values to a signal frequency for data generation. 
It is important to note that the largest eigenvalue of the state-transition matrix
$\mathbf{A}$ 
reflects the frequency of the sine/cosine signal. 

To quantitatively assess the quality of the fitting results, we design a
``denoising'' experiment. The data to be used for denoising is generated
as follows: First, we use each regression method to compute a 
model on the (clean) data points we just generated. In the second
step, we take the initial conditions of each model, shoot forward and 
record the points on the regression curve at fixed values of the independent
variable (\ie, \rk{the signal} frequency). These points serve as our \textit{ground truth}.
In a final step,
we take each point on the ground truth model, generate a random tangent 
vector at that location and shoot forward along that vector for a small time 
(\eg, $0.03$). The newly generated points then \mn{serve} as the ``noisy'' \mn{measurements}
of the original points.

To obtain fitting results for the noisy data, we initialize the first state $\mathbf{X}_1$ 
with the first data point, and all other initial conditions with $\mathbf{0}$.
Table~\ref{tab:synthetic_data_test} lists the differences between our estimated regression curves and the corresponding
ground truth using two strategies: (1) comparison of the initial conditions
as well as the parameters of \rk{the warping function} in TW-GGR; (2) 
comparison of the full curves (sampled at the values of the independent variable) 
and the data points. The numbers indicate that all three models allow us 
to capture different types of relationships on the Grassmannian. 

\begin{figure}[t!]
\centering
\includegraphics[width=0.99\columnwidth]{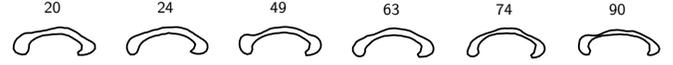}
\caption{\small{\mn{Corpora} callosa (with the subject's age) \cite{Fletcher13geodesic}.}\label{fig:cc_data}}
\end{figure}

\begin{figure}[t!]
\centering
\includegraphics[width=0.95\columnwidth]{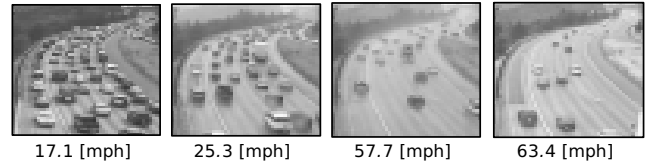}
\caption{\small{Examples of the UCSD traffic dataset \cite{Chan05a} with associated speed measurements.}\label{fig:traffic_data}}
\end{figure}

\begin{figure}[t!]
\centering
\includegraphics[width=0.95\columnwidth]{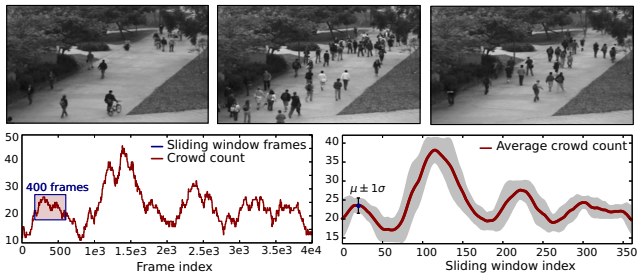}
\caption{\small Example frames from the UCSD pedestrian dataset \cite{Chan12a} (\texttt{Peds1}).
\textit{Bottom}: Total people count over all frames (left), and average people
count over a $400$-frame sliding window (right). \label{fig:crowd_data}}
\end{figure}

\section{Applications}
\label{section:experiments}

To demonstrate Std-GGR, TW-GGR and CS-GGR on 
actual vision data, we present four applications: in the first two 
applications, we regress the manifold-valued variable, \ie,
landmark-based shapes; in the last two applications, we predict 
the independent variable based on the regression curve fitted 
to the manifold-valued data, \ie, LDS representations of 
surveillance videos.

\begin{figure*}
\begin{tikzpicture}[thick, spy using outlines={rectangle,lens={scale=1.5}, width=2.5cm, height=1.2cm, connect spies}]
	\node (reg_id1) {\includegraphics[scale=1,page=1]{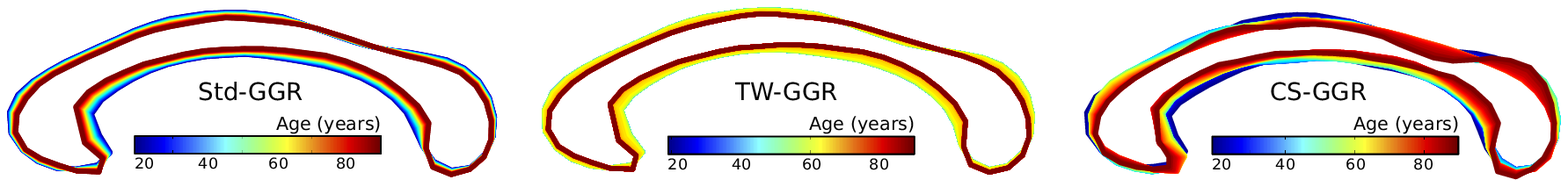}};
 	\spy [black] on (-2.2,-0.6) in node (lspy) [left,fill=black!5] at (1.5,1.8);
	\spy [black] on (-8.35,-0.6) in node (lspy) [left,fill=black!5] at (-4.65,1.8);
	\spy [black] on (4.1,-0.6) in node (lspy) [left,fill=black!5] at (7.5,1.8);	
\end{tikzpicture}
\caption{\small{\label{fig:corpus_callosum}Comparison between 
Std-GGR, TW-GGR and CS-GGR (with one control point) on the 
corpus callosum data \cite{Fletcher13geodesic}. The shapes are 
generated along the 
fitted curves and are colored by age (best viewed in color).}}\end{figure*}

\begin{figure*}[t!]
\begin{tikzpicture}[thick, spy using outlines={rectangle,lens={scale=3.2}, size=1.5cm, connect spies}]
	\node (reg_id1) {\includegraphics[width=0.99\textwidth]{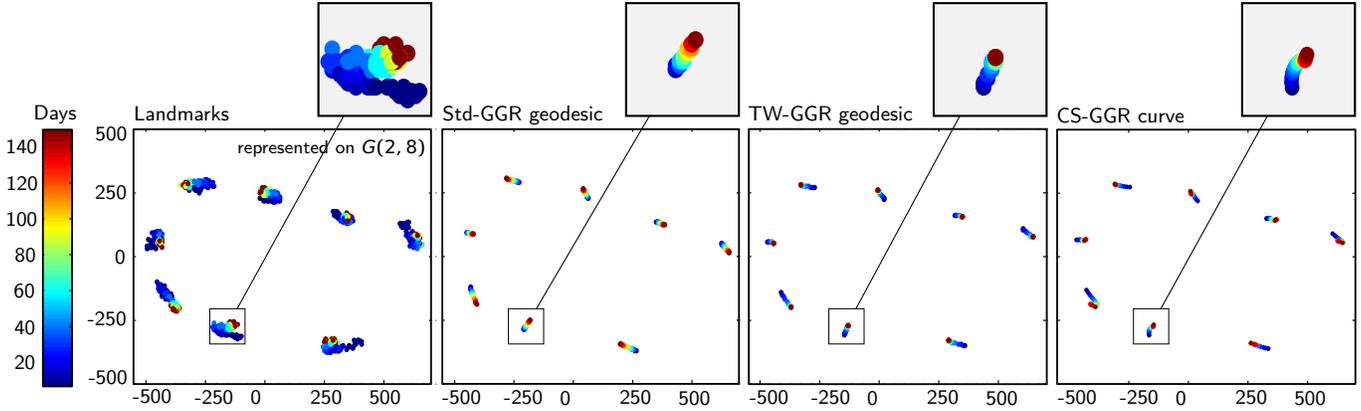}};
	\spy [black] on (-6.03,-0.95) in node (lspy) [left,fill=black!5] at (-3.32,2.6);	
	\spy [black] on (-2.06,-0.95) in node (lspy) [left,fill=black!5] at (0.77,2.6);
	\spy [black] on (2.2,-0.95) in node (lspy) [left,fill=black!5] at (4.86,2.6);
	\spy [black] on (6.25,-0.95) in node (lspy) [left,fill=black!5] at (8.96,2.6);
\end{tikzpicture}
 \caption{\small{\label{fig:rat_calvarium}Comparison between 
Std-GGR, TW-GGR and CS-GGR (with one control point) on the 
rat calvarium data \cite{bookstein1991}. The shapes are generated along the 
fitted curves and the landmarks are colored by age in days (best-viewed in color).} }
\end{figure*}

\subsection{Datasets}


\noindent
\textbf{Corpus callosum shapes~\cite{Fletcher13geodesic}.}
We use a collection of $32$ corpus callosum shapes with ages varying from
19 to 90 years, see Fig.~\ref{fig:cc_data}. Each shape is represented by $64$ 
2D boundary landmarks, and is projected to a point on the Grassmannian using 
the representation \rk{of} Section~\ref{subsubsection:shapes}.

\noindent
\textbf{Rat calvarium landmarks~\cite{bookstein1991}.}
We use 18 individuals with 8 time points from the
Vilmann rat data{\footnote{Online: \url{http://life.bio.sunysb.edu/morph/data/datasets.html}},
each in the age range of 7 to 150
days. Each shape is represented by a set of 8 landmarks. Fig.~\ref{fig:rat_calvarium}~(left)
shows the landmarks projected onto the Grassmannian, using the \rk{same 
representation as the corpus callosum data.}

\noindent
\textbf{UCSD traffic dataset \cite{Chan05a}.} This dataset was introduced in 
the context of clustering traffic flow patterns with LDS models.
It contains a collection of short traffic video clips, acquired by a surveillance system monitoring highway traffic. There are $253$ videos in total and each video is roughly matched to the speed measurements from a highway-mounted speed sensor. We use the pre-processed video clips introduced in~\cite{Chan05a} which were converted to grayscale and spatially normalized to $48 \times 48$ pixels with zero mean and unit variance. Our rationale for using an LDS representation 
for speed prediction is the fact that clustering and categorization experiments in
\cite{Chan05a} showed compelling evidence that dynamics are indicative
of the traffic class. We argue that the notion of speed of an object (\eg, a car) 
could be considered a property that humans infer from its visual dynamics.


\noindent
\textbf{UCSD pedestrian dataset \cite{Chan12a}.} We use the \texttt{Peds1} subset
of the UCSD pedestrian dataset which contains $4000$ frames with a 
ground-truth people count 
associated with each frame.
Fig. \ref{fig:crowd_data} (bottom left) shows the total people count over all frames.
Similar to \cite{Chan12a} we ask the question whether we can infer the number
of people in a scene (or clip) without actually detecting the people. While \rk{this problem has been
addressed} by resorting to crowd~/~motion segmentation and Gaussian process regression
on low-level features extracted from the segmentation regions, we go one step further and 
try to avoid any preprocessing at all. In fact, our objective is to infer an \textit{average} 
people count from an LDS representation of short video segments (\ie, within a 
temporal sliding window). This is plausible because the visual dynamics of a scene 
change as people appear in it. In fact, it could be considered as another form
of ``traffic''. Further, an LDS does not only model the dynamics, but 
also the appearance of videos; both aspects are represented in the observability matrix of the 
system. We remark, though, that such a strategy does not allow for fine-grain 
frame-by-frame predictions as in \cite{Chan12a}. Yet, it has the advantages 
of not requiring any pre-selection of features or potential unstable preprocessing 
steps such as the aforementioned crowd segmentation.

In our setup, we split the $4000$ frames into $37$ video clips of $400$ frames each,
using a sliding window with steps of $100$ frames, illustrated in Fig.~\ref{fig:crowd_data}
(bottom right), and associate an average people count 
with each clip. The video clips are spatially down-sampled to a 
resolution of $60 \times 40$ pixel (original: $238 \times 158$) to keep the 
observability matrices at a reasonable size.
Since the overlap between the clips potentially biases the experiments, we introduce
a weighted variant of system identification (see \mn{supplementary} material) 
with weights based on a Gaussian function centered at the middle of the sliding window and a standard deviation of $100$. While this ensures stable system identification, by still using $400$
frames, it reduces the impact of the overlapping frames 
on the parameter estimates. With this strategy, the average crowd count is 
localized to a smaller region.



\renewcommand{\arraystretch}{1.2}
\begin{table*}[t!]
\begin{center}
\begin{tabular}{|r|cccc|cccc|}
\hline
\multirow{2}{*}{}& \multicolumn{4}{c|}{\bf Corpus callosum \cite{fletcher2013}} & \multicolumn{4}{c|}{\bf Rat calvarium \cite{bookstein1991}} \\
\cline{2-9}
        & Std-GGR & TW-GGR & CS-GGR (1) & CS-GGR (2) & Std-GGR & TW-GGR & CS-GGR (1) & CS-GGR (2)\\
\hline
Energy 	& 0.35    	& 0.34     	&0.32 	&  \cellcolor{green!10}{\bf 0.31} 	& 0.32  	& 0.18    	& \cellcolor{green!10}{\bf 0.16}    & \cellcolor{green!10}{\bf 0.16} \\
$R^2$ 	& 0.12     	& 0.15	& 0.21     	& \cellcolor{green!10}{\bf 0.23}   	& 0.61	& 0.79     	& \cellcolor{green!10}{\bf 0.81}    &  \cellcolor{green!10}{\bf 0.81} \\
MSE    	& 1.25e-2 	& \cellcolor{green!10}{\bf 1.22e-2}  	&  1.36e-2    	&  1.43e-2         	& 2.3e-3    	& 1.3e-3	& \cellcolor{green!10}{\bf 1.2e-3}   & \cellcolor{green!10}{\bf 1.2e-3}\\
\hline
\end{tabular}
\end{center}
\caption{\small{\label{tab:measurement}Comparison of Std-GGR, TW-GGR and CS-GGR with one (1) and two (2) control points on the corpus callosum and rat calvarium datasets. For \textit{Energy} and \textit{MSE} smaller values are better, for $R^2$ larger values are better.}}
\end{table*}

\subsection{Regressing the manifold-valued variable}
\label{subsection:regressing_the_manifold_valued_variable}

The first category of applications leverages the regressed relationship between the
independent variable, \ie, age, and the manifold-valued dependent variable, \ie, shapes.
The objective is to estimate the shape for a given age. We demonstrate Std-GGR, 
TW-GGR and CS-GGR on both corpus callosum and rat calvarium data. 
Three measures are used to quantitatively compare the regression results: (1) 
the regression \textit{energy}, \ie, the data matching error over all observations; 
(2) the $R^2$ statistic on the Grassmannian, which is between 0 and 1, with 1 
indicating a perfect fit and 0 indicating a fit no better than the
Fr\'echet mean (see~\cite{fletcher2013} for more details); and (3) the \textit{mean squared
error (MSE)} on the testing data, reported in a (leave-one-subject-out) crossvalidation (CV) setup. 

In all experiments of this paper, $\sigma$ in the cost function is set to $1$, the initial point 
is set to be the first data point, and \rk{all} other initial conditions are set to zero. For 
the parameter(s) $\boldsymbol{\theta}$ of TW-GGR, 
we fix $\beta, m =1$ so that $M$ is the time of the maximal growth. One control
point is used in CS-GGR for the following two experiments, which is set 
\mn{to the mean age} of the data points.


\noindent
\textbf{Corpus callosum aging.} 
Fig.~\ref{fig:corpus_callosum} shows the corpus callosum shapes\footnote{The shapes
are recovered from the points along the geodesic 
through scaling 
by the mean singular values of the SVD results.} along the fitted curves for the time points
in the data. Table~\ref{tab:measurement} lists the quantitative measurements. 
With Std-GGR, the corpus callosum 
starts to shrink from age 19, which is consistent with the regression results 
in~\cite{fletcher2013} and \cite{hinkle2014}. However, according 
to biological studies~\cite{Hopper94a,Johnson94a}, the corpus callosum size remains stable 
during the most active years of the lifespan, which is consistent with our TW-GGR result. 
As we can see from the 
optimized logistic function in 
Fig.~\ref{fig:estimated_tw} (left), TW-GGR estimates that thinning starts at  
$\approx50$ years, and at the age of $65$, the shrinking rate reaches its peak. From the CS-GGR results,
we first observe that the $R^2$ value increases notably to $0.21/0.23$, compared 
to $0.12$ for Std-GGR. While this suggests a better fit to the data, it is not a fair
comparison, since the number of parameters for CS-GGR increases as well and
a higher $R^2$ value is expected. Secondly, the more interesting observation 
is that, qualitatively, we observe higher-order shape changes in the anterior 
and posterior regions of the corpus callosum, shown in the 
zoomed-in regions of
Fig. \ref{fig:corpus_callosum}; this is similar to what is reported in \cite{hinkle2014} for
polynomial regression in 2D Kendall shape space. However,
our shape representation on $G(p,n)$, by design, easily extends 
to point configurations in $\mathbb{R}^3$. This is in contrast to 3D Kendall shape 
space which has a substantially more complex structure than its 2D variant \cite{Dryden98a}.

\begin{figure}
\centering
\includegraphics[width=0.85\columnwidth]{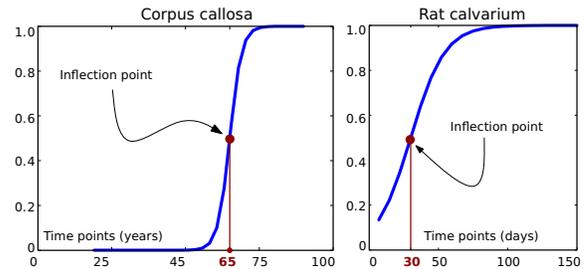}
\caption{\small{Estimated time-warp functions for TW-GGR.}\label{fig:estimated_tw}}
\end{figure}

\noindent
\textbf{Rat calvarium growth.} 
Fig.~\ref{fig:rat_calvarium}~(leftmost) shows the projection of the original data on $G(2,8)$, 
as well as data samples generated along the fitted curves. Table \ref{tab:measurement}
lists the performance measures. From the zoomed-in \mn{regions} in Fig.~\ref{fig:rat_calvarium}, 
we observe that the rat calvarium grows at an approximately 
constant speed during the first 150 days \mn{if} the relationship is modeled by Std-GGR.
However, the estimated logistic curve for TW-GGR, shown in Fig.~\ref{fig:estimated_tw} (right), 
indicates that the rat calvarium only grows fast in the first few weeks, reaching its peak at $~30$ 
days; then, the rate of growth gradually levels off and becomes steady after around $14$ weeks. In fact, 
similar growth curves for the rat skull were reported in~\cite{hughes78}. Based on their study, 
the growth velocities of viscerocranium length and nurocranium width rose to the peak in the 
$26-32$ days period. Comparing the $R^2$ values for TW-GGR and CS-GGR, we see an 
interesting effect: although, we have more parameters in CS-GGR, the
$R^2$ score only marginally improves. This indicates that TW-GGR already sufficiently 
captures the relationship between age and shape. \mn{It} further confirms, to a large
extent, the hypothesis put forward in \cite{hinkle2014}, where the authors noted that 
the cubic polynomial in 2D Kendall shape space essentially degrades to a geodesic
under polynomial time reparametrization. Since TW-GGR, by design, reparametrizes
time (not via a cubic polynomial, but via a logistic function), it is not surprising that 
this relatively simple model exhibits similar performance to the more complex CS-GGR model. 


\renewcommand{\arraystretch}{1.2}
\begin{table*}[t!]
\begin{center}
\begin{tabular}{|r|cccc|cccc|}
\hline
\multirow{2}{*}{} & \multicolumn{4}{c|}{\bf Traffic speed} & \multicolumn{4}{c|}{\bf People counting} \\
\cline{2-9}
 & Baseline & Std-GGR &  Std-GGR (PW) & CS-GGR  & Baseline & Std-GGR & Std-GGR (PW) & CS-GGR \\
\hline
Mean energy &  -- 	& $2554.88$ & \cellcolor{green!10}{$\mathbf{2461.95}$} & $2670.84$ &  -- & $273.81$ & \cellcolor{green!10}{$\mathbf{224.87}$} &$244.02$\\
Train-MAE & 	--	& $2.98\pm0.33$ & \cellcolor{green!10}{$\mathbf{1.48\pm0.07}$} & $2.42\pm0.35$ & --  & $0.97\pm0.07$ & \cellcolor{green!10}{$\mathbf{0.59\pm0.13}$} & $0.63\pm0.19$\\
Test-MAE &  $4.14\pm0.36$  & $4.44\pm0.16$ & \cellcolor{green!10}{$\mathbf{3.46\pm0.64}$} & $6.32\pm1.62$ &$2.40\pm0.53$ & \cellcolor{green!10}{$\mathbf{1.88\pm0.75}$} & $2.14\pm1.03$ &  $2.11\pm0.76$\\
\hline
\end{tabular}
\end{center}
\caption{Mean energy and mean absolute errors (MAE) over all CV-folds $\pm 1\sigma$ on training and testing data.}
\label{tab:measurements_traffic_crowd}
\end{table*}

\noindent
\textbf{Comparison to a Jacobi field approach.} We compare the performance 
of our Std-GGR approach to a comparable approach using Jacobi fields \cite{quentin2011}. 
Both methods are applied on the corpus callosum dataset.
To quantitatively measure 
the differences between the regression results, we first compute the distance between the 
estimated initial conditions. The geodesic distance between two initial points ($\mathbf{X}_1$)
is 9e-4, and the Frobenius norm between the two initial velocities ($\mathbf{X}_2$) is 4e-3. 
Secondly, we use the initial conditions to shoot forward and 
calculate the geodesic distance between corresponding points on the two estimated geodesics. 
The mean geodesic distance is 4e-4, 
which indicates
that both methods provide similar solutions. 
Although they have comparable performance for standard regression, our Std-GGR
\mn{has the advantage of being easily extensible} to higher-order models, \eg, CS-GGR.


\begin{figure}[t!]
	\centering
	\includegraphics[width=0.95\columnwidth]{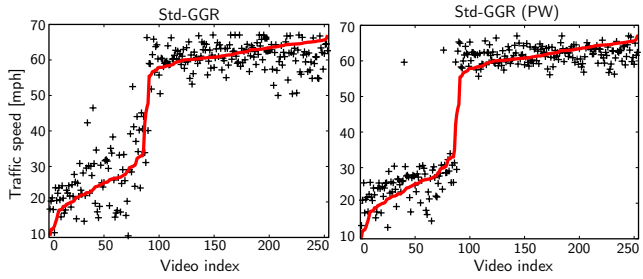}
  \caption{\label{fig:traffic_speed_prediction}\small{Traffic 
  speed predictions via 5-fold CV using Std-GGR (left) and its piecewise variant (right). The red solid curve shows the
ground truth (best-viewed in color).}}
\end{figure}

\vspace{-0.2cm}
\subsection{Predicting the independent variable}

The second category of applications aims to predict the independent variable using its 
regressed relationship with the manifold-valued dependent variable. Specifically, given 
a point on the Grassmannian, \eg, an LDS representation of a video clip, we \textit{search along the
regressed curve} (with a step size of $0.05$ in our experiments) to find its closest point, and 
then take the corresponding independent variable of this closest point as its predicted value. 
This could be considered a variant of nearest-neighbor regression where the search space
\rk{is restricted} to the sampled curve on the Grassmannian. \rk{The case when the search
space is \textit{not} restricted, but contains all points, will be referred to as our
\textit{baseline}. Note that in our strategy, search complexity is controlled via the step-size, 
while the search complexity for the \textit{baseline} scales linearly (for each prediction) 
with the sample size.}

Furthermore, we remark that in this category of applications, TW-GGR is not appropriate for predicting the independent 
variable for the following reasons: First, in case of \mn{the} traffic speed measurement, the generalized 
logistic function tends to degenerate to almost a step-function, due to the limited number of 
measurement points in the central regions. In other words, two greatly different independent 
variables would correspond to two very close data points, even the same one, which would 
result in a large prediction error. Second, in case of crowd-counting, there is absolutely no
prior knowledge about any saturation or growth effect which could be modeled via a logistic
function. Consequently, we only demonstrate Std-GGR and CS-GGR on the two datasets. 
\yi{Note that prediction based on nearest neighbors could be problematic in case of CS-GGR, 
since the model does not guarantee a monotonic curve.}
We report the mean regression \textit{energy} and the \textit{
mean absolute error (MAE)}, 
computed over all folds in a crossvalidation setup with a dataset-dependent number 
of folds. 


\begin{figure}[t!]
	\centering
	\includegraphics[width=0.95\columnwidth]{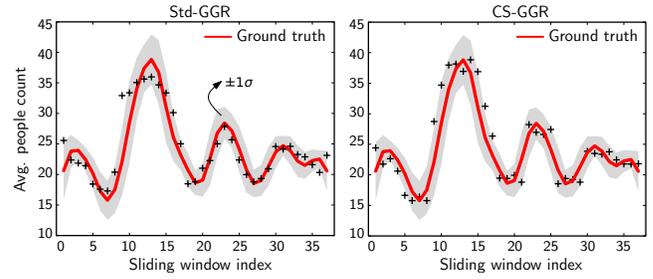}
  \caption{\label{fig:crowd_counting}\small{Crowd counting results 
  via 4-fold CV.
  Predictions are 
  shown as a function of the sliding window index.
  The gray envelope indicates the weighted 
standard deviation ($\pm 1\sigma$) around the average crowd size 
in a sliding window (best-viewed in color).}}
\end{figure}

\noindent
\textbf{Speed prediction.}
For each video clip, we estimate LDS models with $p=10$ states. 
The control point of CS-GGR and the breakpoint for piecewise Std-GGR 
is set at 50 [mph]. Results are reported for 
5-fold CV, see Fig.~\ref{fig:traffic_speed_prediction}.
\yi{The quantitative 
comparison to the baseline in Table~\ref{tab:measurements_traffic_crowd} shows that
piecewise Std-GGR has the best performance.}

\noindent
\textbf{Crowd counting.} For each of the $37$ video clips we extract from the \texttt{Peds1}
dataset, we estimate LDS models with $p=10$ states using weighted system identification.
For CS-GGR, the control point is set 
to a count of 23 people which separates the 37 videos into two groups of roughly 
equal size. Quantitative results for 4-fold CV are reported in Table~\ref{tab:measurements_traffic_crowd}, 
Fig.~\ref{fig:crowd_counting} shows the predictions \textit{vs.} the ground truth. As we see, both 
Std-GGR and CS-GGR output predictions ``close'' to the ground truth, mostly within 
$1\sigma$ (shaded region) of the average crowd count. 
However, a closer look at  Table~\ref{tab:measurements_traffic_crowd} reveals a 
typical overfitting effect for CS-GGR: while the training MAE is quite low, the
testing MAE is higher than for the simpler Std-GGR approach. While both models 
exhibit comparable performance (considering the standard deviation of $\approx 0.7$), 
Std-GGR is preferable, due to fewer parameters \yi{and its guaranteed 
monotonic regression curve.}

\vspace{-0.05cm}
\section{Discussion}
\label{section:conclusion}

In this paper, we developed a general theory for parametric regression 
on the Grassmann manifold from an optimal-control perspective. By 
introducing the basic principles for fitting models of increasing 
\mn{order} for the special case of $\mathcal{M} = \mathbb{R}^n$, we established 
the framework that was then used for a generalization to Riemannian 
and, in particular, the Grassmann manifold. We demonstrated that our 
solution to the parametric regression problem is simple, extensible, 
and easy to implement. 

From an application point of view, we have seen that quite different 
vision problems can be solved within the same framework under minimal 
data preprocessing. While the presented applications are limited to 
shape analysis and surveillance video processing, our method should, 
in principle, be widely applicable to other problems on the Grassmann 
manifold, \eg, domain adaptation, facial pose regression, or the 
recently proposed domain evolution problems.

Regarding the limitations of the proposed approach, we note that the issue of 
model selection is critical. In fact, whether we should use Std-GGR, TW-GGR
or CS-GGR highly depends on our prior knowledge of the data. 
In shape regression, for instance
\mn{such prior knowledge is frequently available,}
since the medical~/~biological literature already provides evidence for different 
growth and saturation effects as a function of age. For applications 
where prediction of the independent variable is of importance, \eg, traffic or
or crowd surveillance, we additionally have computational constraints in many
cases. Interestingly, a simple geodesic curve as a model for regression can 
often provide sufficiently good performance, as we observed in the crowd counting
experiment. We hypothesize that this can be explained, to some 
extent, by the fact that geodesic regression respects the geometry of 
the underlying space. It is possible that in this space, the relationship between
the dependent and the independent variable might actually be relatively simple 
to model. In contrast, approaches where video content is boiled down to 
feature vectors and conventional regression approaches with standard kernels 
are used, more flexible models might be needed. TW-GGR can serve as a hybrid
solution when we have prior knowledge about the data; 
however, samples throughout
the range of the independent variable are needed to avoid degenerate cases 
of the warping function. While this could be avoided via regularization, we did not
explore this direction.


\appendices

\ifCLASSOPTIONcaptionsoff
  \newpage
\fi



\bibliographystyle{IEEEtran}
\bibliography{IEEEabrv,egbib}
\end{document}